%% file: main.tex
\documentclass[conference]{IEEEtran}
\usepackage{cite}
\usepackage{amsmath,amssymb,amsfonts}
\usepackage{algorithmic}
\usepackage{graphicx}
\usepackage{textcomp}
\usepackage{xcolor}
\usepackage{subfig}
\usepackage{multicol}

\usepackage[belowskip=0pt,aboveskip=0pt]{caption}

\usepackage{hyperref}   
\usepackage{setspace}

\ifCLASSINFOpdf
\else
\fi
\begin{document}
\makeatletter 
\def\@IEEEpubidpullup{8\baselineskip} 
\makeatother 
\IEEEoverridecommandlockouts
\IEEEpubid{
\parbox{\columnwidth}{\vspace{-4\baselineskip} Permission to make digital or hard copies of all or
part of this work for personal or classroom use is granted without fee provided that copies are not
made or distributed for profit or commercial advantage and that copies bear this notice and the full
citation on the first page. Copyrights for components of this work owned by others than ACM must
be honored. Abstracting with credit is permitted. To copy otherwise, or republish, to post on servers
or to redistribute to lists, requires prior specific permission and/or a fee. Request permissions from
\href{mailto:permissions@acm.org}{permissions@acm.org}.\hfill\vspace{-0.8\baselineskip}\\
\begin{spacing}{1.2}
\small\textit{ASONAM '21}, November 8-11, 2021, Virtual Event, Netherlands \\
\copyright\space2021 Association for Computing Machinery. \\
ACM ISBN 978-1-4503-9128-3/21/11\$15.00 \\
\url{http://dx.doi.org/10.1145/3487351.3488276}
\end{spacing}
\hfill}
\hspace{0.9\columnsep}\makebox[\columnwidth]{\hfill}}
%\IEEEpubidadjco

%
% paper title
% Titles are generally capitalized except for words such as a, an, and, as,
% at, but, by, for, in, nor, of, on, or, the, to and up, which are usually
% not capitalized unless they are the first or last word of the title.
% Linebreaks \\ can be used within to get better formatting as desired.
% Do not put math or special symbols in the title.
\title{Will You Dance To The Challenge? \\Predicting User Participation of TikTok Challenges}

% author names and affiliations
% use a multiple column layout for up to three different
% affiliations
%\author{\IEEEauthorblockN{Michael Shell}
%\IEEEauthorblockA{School of Electrical and\\Computer %Engineering\\
%Georgia Institute of Technology\\
%Atlanta, Georgia 30332--0250\\
%Email: http://www.michaelshell.org/contact.html}
%\and
%\IEEEauthorblockN{Homer Simpson}
%\IEEEauthorblockA{Twentieth Century Fox\\
%Springfield, USA\\
%Email: homer@thesimpsons.com}
%\and
%\IEEEauthorblockN{James Kirk\\ and Montgomery Scott}
%\IEEEauthorblockA{Starfleet Academy\\
%San Francisco, California 96678--2391\\
%Telephone: (800) 555--1212\\
%Fax: (888) 555--1212}}

\author{\IEEEauthorblockN{Lynnette Hui Xian Ng\textsuperscript{1}, John Yeh Han Tan\textsuperscript{2}, Darryl Jing Heng Tan\textsuperscript{3}, Roy Ka-Wei Lee\textsuperscript{3}}
\IEEEauthorblockA{\textsuperscript{1}Carnegie Mellon University, \textsuperscript{2}National University of Singapore, \textsuperscript{3}Singapore University of Technology and Design \\
Email: lynnetteng@cmu.edu, johntyh@u.nus.edu, \{darryl\_tan, roy\_lee\}@sutd.edu.sg}
}

% make the title area
\maketitle

\begin{abstract}
TikTok is a popular new social media, where users express themselves through short video clips. A common form of interaction on the platform is participating in ``challenges", which are songs and dances for users to iterate upon. Challenge contagion can be measured through replication reach, i.e., users uploading videos of their participation in the challenges. The uniqueness of the TikTok platform where both challenge content and user preferences are evolving requires the combination of challenge and user representation. 
This paper investigates social contagion of TikTok challenges through predicting a user's participation. We propose a novel deep learning model, \textsf{deepChallenger}, to learn and combine latent user and challenge representations from past videos to perform this user-challenge prediction task. We collect a dataset of over 7,000 videos from 12 trending challenges on the ForYouPage, the app's landing page, and over 10,000 videos from 1303 users. Extensive experiments are conducted and the results show that our proposed \textsf{deepChallenger} (F1=0.494) outperforms baselines (F1=0.188) in the prediction task.
\end{abstract}


\section{Introduction}
TikTok is a new video-based social media platform created by the Chinese company ByteDance. It is increasingly population among the younger demographic, with \~60\% of its users below the age of 34 \cite{tankovska2021} in July 2020. Its content features short video clips of under a minute. Upon opening the application, the app lands on the ForYouPage, which aggregates top content in the user's geographical region with recommended content based on the user's interaction (likes, shares) \cite{tiktok2019}. 
TikTok Challenges are video formats that users endlessly iterate upon, usually involving an easy to replicate song and dance. Popular challenges in a user's region appear in the ForYouPage to encourage users to join in with their fellow TikTokers. 

\textbf{Research Objectives.} In this work, we predict the contagion of 12 TikTok Challenges that trend on the ForYouPage during the COVID19 lockdown. This work draws ideas from collaborative filtering, which exploit the binary similarities between users and items to recommend content to users \cite{sarwar2001item, 8618448}. Research efforts in this area utilize video content for recommendation on platforms like YouTube \cite{davidson2010youtube, baluja2008video} and enhance them with user feedback such as click-through rate and ratings \cite{mei2011contextual, yang2007}. In this study, we extend the problem from predicting the viewership of a video to predicting user participation in a challenge. Challenge video content are not static as each user injects his own flavour into the challenge; performing a challenge is thus a combination of user's interests and the challenge topics. 

We combine a user's previously uploaded videos and videos uploaded on the challenge to perform to predict whether a user will catch on the challenge contagion and participate in a challenge. With a total of 10,424 TikTok videos from 1303 users, we construct a \textsf{deepChallenger} model, a novel deep learning framework, which combines learned embeddings of challenge and user latent representations to perform this user-challenge participation predictive task. Our results show that the \textsf{deepChallenger} model outperforms (F1=0.494) the baseline models (F1=0.188).

\section{Background}
As of August 2020, TikTok has about 100 million monthly active users \cite{sherman4949_2020}. Its content composes of either 15s or 60s short video clips. With the merger of music app Musical.ly \cite{russell_2018}, TikTok became more sound centric, allowing users to pick background sounds to match to their videos. Users create content by uploading a video to the application, along with adding video filters, background music and captions. Captions usually include popular hashtags that helps the video to get noticed through hashtag searches. Although TikTok has a following-follower structure, content is mainly consumed through the aggregated ForYouPage, rather than content solely produced by those a user is following.

Since the inception of the platform, studies have examined the causes of popularity of this new social media platform \cite{xu2019research} and motivations of the young TikTok user demographic \cite{ahlse2020s}. Other work examined communication of partisan views through a platform-unique feature called TikTok duets \cite{10.1145/3394231.3397916}.

TikTok challenges are another platform-unique feature, designed with a central dance or action to be replicated by other users and are branded with a challenge hashtag. They gain traction by inspiring their audiences to participate in them. The most popular challenges propagate widely through the app, reaching audiences across geographical boundaries. The \#washyourhands challenge is a simple dance promoting washing of hands, aiming to reduce the spread of COVID19; \#attentionchallenge shows off gymnastics moves done at home due to the unavailability of community gyms during lockdown. 

In the examining content contagion, previous work predicted popularity of video posts shared on social media platforms \cite{cui2011should, ding2015video} and hashtags given to Twitter posts \cite{yang2012we, ben2018using}. However, studies that examine user-video content are few, and are mostly survey-based studies \cite{MYRICK2015168}. 
Spread of ideas in TikTok differ from commonly studied social media platforms like Twitter. In the absence of platform-defined distribution processes like retweet dynamics as a measure of social contagion \cite{ICWSM1613026,doi:10.1126/science.aap9559}, we measure social contagion through user participation rate. As users participate in a challenge, the higher chance it appears in another user's ForYouPage, which creates a cycle of participation.
This study fills the research gap by investigating and predicting user's participation in TikTok challenges. 

\section{Dataset}

We collected challenges and users videos data from TikTok using the Python API wrapper by davidteather\footnote{https://github.com/davidteather/TikTok-Api}. This wrapper requires web credentials from a valid TikTok account. 
While conducting this study, some ethical issues must be taken into consideration. When a user posts a TikTok video, he can select one of the three audience types: everyone, friends only or himself. We only collected videos publicly available for everyone. In order to maintain user privacy, no user identity data was collected nor used in the analysis; a unique user identifier was used to correlate user videos.

\subsection{Challenge data collection}
We collected videos from 12 trending TikTok challenges that were initiated from 19 October to 19 December 2020, reflecting health campaigns and life surrounding the coronavirus pandemic. We created a TikTok account from Singapore and collected the top challenges from the ForYouPage. Challenges on the ForYouPage are recommended to users around the region for its engagement metrics, one of which is the number of users that watch the video till its end \cite{feldman_2019}. The challenges trending on the account's ForYouPage has two key categories: (1) challenges that aim to promote awareness on good hygiene habits to reduce the spread of the virus (eg \#washyourhands); and (2) challenges that provide entertainment through dance and song during the quarantine period that governments enforced to control the pandemic (eg \#BoredInTheHouse). These videos have a view count of 0.5 to over 2 million. We collected the following challenges: \textit{\#attentionchallenge, \#blindinglightschallenge, \#100wayschallenge, \#BoredInTheHouse, \#karenchallenge, \#Cannibalchallenge, \#renegadechallenge, \#washyourhandschallenge, \#ghencovychallenge, \#godaddygo, \#papertowelchallenge}.

We first searched for videos related to the challenge using the challenge hashtag, then downloaded the video by its ID. We collected around 1000 videos from each challenge, except when the crawler did not return the desired number of videos, which we posit could be due to the crawler limitations and TikTok restrictions on the search query duration.

\subsection{User data collection}
Due to data collection constraints, we were unable to collect all the 7075 users present in the dataset. We thus collected the most representative users. We first analyzed the number of challenges each user performed in our Challenge dataset and collected users that performed at least 3 of our 12 challenges. We first searched for videos the user uploaded, then downloaded the video mp4 file. We excluded videos that contained the hashtag of the challenges we are focusing on. We ran this collection mechanism for a month, and collected TikToks from 1303 users, with 8 videos each. In total, we collected 10,424 videos from the selected users. 

\section{Proposed Model}
In this section, we elaborate on our proposed \textsf{deepChallenger} framework. The intuition of our proposed framework is to learn the latent representations of the users and challenges using two proxy tasks, and effectively combine the learned embeddings to predict user's involvement in TikTok challenges. In the subsequent sections, we describe the two main components of our framework:  (a) challenge representation learning, and (b) user representation learning. Finally, we present how the learned representations are used to perform the user-challenge participation classification task. 

\begin{figure*}[t]
  \centering
  \includegraphics[width=0.8\textwidth]{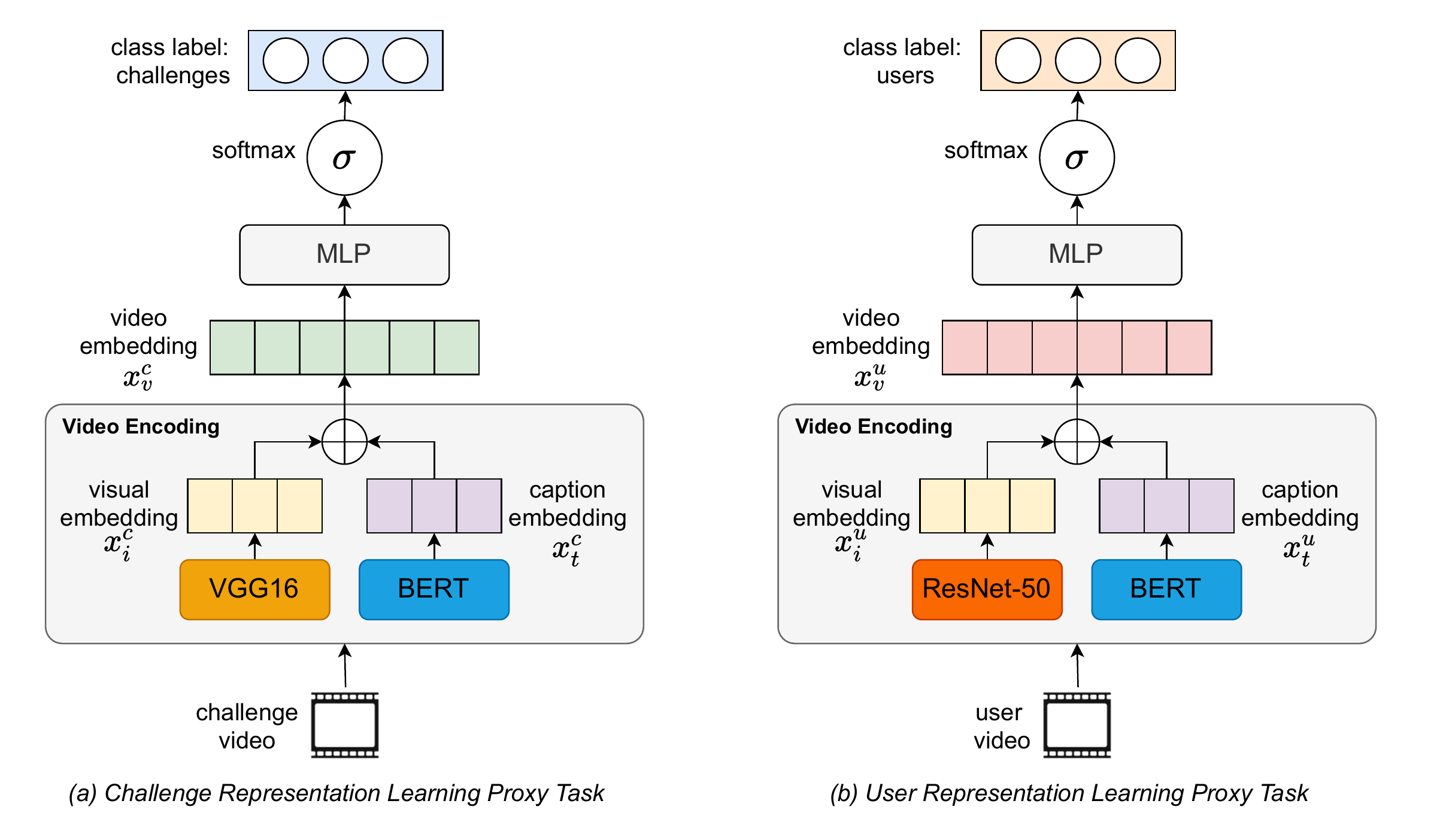}
  \caption{Proxy tasks for (a) challenge and (b) user representation learning}
  \label{fig:proxy}
\end{figure*}

\subsection{Challenge Representations Learning}
To learn the latent representation of challenges, we first formulate a proxy task: ``\textit{Given a TikTok video, predict which challenge it belongs to.}''  Figure \ref{fig:proxy}(a) illustrates our proposed approach to perform this proxy task, in which we encode an input challenge video $v$. Specifically, the video encoding models two aspects of $v$: the visual sequential image frames and the textual caption. 

For the visual sequential image frames, we first breakdown the video into 50 image frames and input each image frame into a pre-trained VGG16 \cite{simonyan2015deep} model to extract the frame's embedding $e^i_{f}$. Next, we concatenate all the image frames' embeddings to form a visual embedding of the video, i.e., $x^c_{i} = (e^i_{f1} \oplus e^i_{f2} \cdots \oplus e^i_{f50})$. Each image frame is pre-processed by resizing to (224,224) and normalized.

To model the video's textual caption, we first tokenize the caption using default tokenizer of BERT language model \cite{devlin2019bert} and extract the pre-trained BERT embeddings: $E_B=\{e^b_1,e^b_2,\dots,e^b_n\}$, $e^b_i\in R^{768}$. A pre-trained BERT model recieves these embeddings as an input, and outputs the \textsf{[CLS]} token caption embedding as $x^c_{t}$. 

Finally, we represent a video embedding as the concatenation of its visual embedding and caption embedding, i.e., $x^c_{v}= (x^c_{i} \oplus x^c_{t})$. The video embedding $x^c_{v}$ is then fed into a feed forward layer to predict the video's challenge class label.

After training the proposed model in Figure \ref{fig:proxy}(a) with the proxy task, we learn a set of video embeddings that are predictive of the challenge videos. The underlying intuition is that learned video embeddings contain latent attributes that are descriptive of the challenges. Therefore, for each challenge, we denote the challenge's latent presentation as:

\begin{equation}
    x^c = (x^c_{v1} \oplus x^c_{v2}\cdots \oplus x^c_{vn})
\end{equation}

where $x^c$ is a concatenation of the learned video embeddings of the challenge-related videos, and $n$ is the number of videos that belong to the challenge. The challenge representation will be used in the user-challenge participation classification task described in Section \ref{sec:userchallengeparticipation}.

\subsection{User Representations Learning}
We adopt a similar approach to learn the user's latent presentation. We formulate a proxy task: ``\textit{Given a video, predict which user it belongs to.}'' Figure \ref{fig:proxy}(b) illustrates our proposed approach to perform the proxy task, where we first encode an input user video $v$ by modeling its visual sequential image frames and textual caption.

Differing from the video encoding used in challenge representation learning, we adopt a pre-trained ResNet-50 \cite{he2015deep} to model the image frames and extracted the image frame's embedding $e^i_{f}$. ResNet-50 is preferred over VGG16 because through empirical experiments, ResNet-50 yields better results for this proxy task. Next, we concatenate all the image frames' embeddings to form a visual embedding of the video, i.e., $x^u_{i} = (e^i_{f1} \oplus e^i_{f2} \cdots \oplus e^i_{f50})$. 

The video's textual caption is modeled using the same approach as described in earlier section. We use the output from the BERT's \textsf{[CLS]} token caption embedding as $x^u_{t}$. 

The user's video embeddings are represented as concatenation of it's visual embedding and caption embedding, i.e., $x^u_{v}= (x^u_{i} \oplus x^u_{t})$. The video embedding $x^u_{v}$ is input into a feed forward layer to predict the user class label of the video.

Using the same intuition for the challenge latent representation learning, the video embeddings learned with the user representation learning proxy task would contain latent attributes that are descriptive of the users. Therefore, for each user, we denote the user's latent presentation as:

\begin{equation}
    x^u = (x^u_{v1} \oplus x^u_{v2}\cdots \oplus x^u_{vm})
\end{equation}

where $x^u$ is a concatenation of the learned video embeddings of the videos posted by the users, and $m$ is the number of videos from the users. The user representation will be used in the user-challenge participation classification task described in Section \ref{sec:userchallengeparticipation}.

\subsection{User-Challenge Participation}
\label{sec:userchallengeparticipation}
%Figure \ref{fig:main} illustrates the user-challenge participation task in our \textsf{deepChallenger} framework. 
Finally, we perform the binary classification task: ``\textit{Given a user and a challenge, predict whether the user will participate in the challenge.}'' The challenge embedding $x^c$ and user embedding $x^v$ are fed into a feed-forward layer for user-challenge participation prediction. Our model's intuition is that the challenge and user embeddings would capture latent features that are predictive of the user's preferences towards challenge participation. We note that user-posted videos that belongs to the challenge video dataset are excluded when learning the user representation. This prevents bias from using overlapping training data to model both challenge and user representations.

\subsection{Implementation Details}
We used the same embedding size for pre-trained BERT embeddings and the same number of video frames $f=50$ for video embeddings. Additionally, we added a dropout layer with 20\% dropout for the fully connected layer. We use the ADAM optimizer \cite{kingma2017adam} with a learning rate of 0.001 to train our models. We used a batch size 4 to optimise for the large model size, with 30 epochs of training optimised for early stopping and a three-fold cross-validation.

\section{Experiments}
\subsection{Experimental Settings}

\textbf{Baselines.} As we are proposing a novel prediction task, there are no known baseline that predicts user-challenge participation. Therefore we designed the following baselines that utilized features of user's posted videos to predict the user's challenge participation:

\begin{itemize}
    \item \textit{VGG16}: We utilize the visual embeddings of user's videos to train a deep neural network (DNN) classifier to predict if the user would would participate in a given challenge. The visual embeddings are extracted using a pre-trained \textit{VGG16}~\cite{simonyan2015deep}.
    \item \textit{VGG16+BERT}: \textit{VGG16} baseline enhanced with caption embeddings learned using pre-trained BERT \cite{devlin2019bert}. 
    \item \textit{ResNet-50}: We adopt the similar approach used in \textit{VGG16} baseline but adopted the pre-trained \textit{ResNet-50} \cite{he2015deep} to extract the visual embeddings of the user's videos.
    \item \textit{ResNet-50+BERT}: \textit{ResNet-50} baseline enhanced with caption embeddings learned using pre-trained BERT.
\end{itemize}

\textbf{Training and Testing Set}. In our experiments, we adopt an 80-20 split, where for each experimental run, 80\% of the videos are used for training with the remaining 20\% used for testing. This setting also applies to the challenge and user latent representation learning proxy tasks.

\textbf{Evaluation Metrics}. We use the macro averaging precision (Macro-F1), recall (Macro-Rec) and F1 score (Macro-F1) as the evaluation metrics. Three-fold cross-validation is used in our experiments and the average results are reported.
\subsection{Experimental Results}
\input{proxy_result}
\input{main_result}
In Table \ref{tab:proxy}, we report the performance on the proxy tasks. We observed that models that utilized textual caption information yield better performance on both proxy tasks. Specifically, we observe that the VGG16+BERT model performs better for the challenge representation learning task, while ResNet-50+BERT achieves better performance for the user representation learning task. Motivated by these empirical results, we adopt the superior model configurations to learn the challenge and user representations in our \textsf{deepChallenger} framework.

Table \ref{tab:main} shows the experimental results for the user-challenge participation prediction task. We observe that \textsf{deepChallenger} outperforms the best baseline (F1=0.494 vs F1=0.188). We note that the baseline models with caption embeddings (i.e., models with BERT) outperforms the baseline models that utilized only visual embeddings (i.e., VGG16 and ResNet-50). Nevertheless, the \textit{VGG16+BERT} and \textit{ResNet-50+BERT} baselines are observed to achieve similar performance. 
The main difference between \textsf{deepChallenger} and the baselines is that \textsf{deepChallenger} learned and utilized the challenge embeddings for the final prediction task. Thus, the superior performance of \textsf{deepChallenger} underscores the importance of incorporating challenge representations in predicting user challenge adoption.

\section{Discussion and Conclusion}
In this study, we study social contagion of TikTok challenges by predicting whether users will participate in a challenge, given their past video uploads and the challenge videos. Both factors are continually evolving across time. We used the top 1000 trending challenge videos to characterize challenges as these videos are most likely to appear on many user's ForYouPage. We use the latest 8 videos of each user to characterize a user's interests.

Our proposed \textsf{deepChallenger} model performs 2.5 times better (F1=0.494) than the user-challenge participation baselines, which only uses user past data to predict whether a user will participate in the challenge. We posit that the low F1 score for the baseline task (F1=0.188) could be due to the data sparsity: there is an overwhelming number of users compared to the number of videos each user is represented by.

The user representation learning task characterizes what type of video a user will upload, while the challenge representation learning proxy task characterizes the types of videos that are common of each challenge. In investigating whether a user will adopt a certain social challenge, it is not sufficient to look only at a user's past videos (F1=0.188), but we must look at the nature of the challenge videos. A user is most likely to participate in a challenge if the challenge nature suits his interests. The increase in model performance upon adding challenge representation learning shows that challenge actions must match user's interests (gleaned from past videos) for a user to catch the challenge virus.

As with any study, there are several limitations of this work. Since TikTok does not have a native API for data collection, we used a wrapper available on Github, which is highly dependent on the website structure. As a fast-developing platform, TikTok's webpage structure constantly changes, during which data collection is impacted and we had to patch the API before resuming collection. %Further, the challenges in our account's ForYouPage is largely region-specific, and so are the top TikToks that are returned for each challenge, and future work includes creating accounts from different regions to gather a more complete result set. 

Our study on TikTok social contagion is part of a broader topic of user engagement on social media. TikTok presents a new medium for viral content adaptation, namely through video for content spread.  For future work, we hope to characterize the propagation and replication reach of ForYouPage challenges, dances and songs on this wildly popular platform. 
%We will also explore more multimodal deep learning methods to perform the user-challenge participation prediction task. 

% trigger a \newpage just before the given reference
% number - used to balance the columns on the last page
% adjust value as needed - may need to be readjusted if
% the document is modified later
%\IEEEtriggeratref{8}
% The "triggered" command can be changed if desired:
%\IEEEtriggercmd{\enlargethispage{-5in}}

% references section

% can use a bibliography generated by BibTeX as a .bbl file
% BibTeX documentation can be easily obtained at:
% http://mirror.ctan.org/biblio/bibtex/contrib/doc/
% The IEEEtran BibTeX style support page is at:
% http://www.michaelshell.org/tex/ieeetran/bibtex/
\bibliographystyle{IEEEtran}
% argument is your BibTeX string definitions and bibliography database(s)
%\bibliography{IEEEabrv,../bib/paper}
%
% <OR> manually copy in the resultant .bbl file
% set second argument of \begin to the number of references
% (used to reserve space for the reference number labels box)
%\begin{thebibliography}{1}
\bibliography{main}
%\bibitem{IEEEhowto:kopka}
%H.~Kopka and P.~W. Daly, \emph{A Guide to \LaTeX}, 3rd~ed.\hskip 1em plus
%  0.5em minus 0.4em\relax Harlow, England: Addison-Wesley, 1999.

%\end{thebibliography}

% that's all folks
\end{document}

%% file: proxy_result.tex
\begin{table}[t]
\caption{Performance on challenge and user representation learning proxy tasks}
\label{tab:proxy}
\begin{center}
% \small
%  \resizebox{0.5\textwidth}{!}{
\begin{tabular}{c|ccc}
\hline
\textbf{Model} & \textbf{Macro-Prec} & \textbf{Macro-Rec} & \textbf{Macro-F1} \\ \hline \hline
\multicolumn{4}{l}{\textbf{Challenge Representation Learning}} \\ \hline
VGG16 & 0.248 & 0.186 & 0.153 \\ 
VGG16 + BERT & \textbf{0.660} & \textbf{0.563} & \textbf{0.494} \\ 
ResNet-50 & 0.257 & 0.174 & 0.125 \\ 
ResNet-50 + BERT & 0.583 & 0.486 & 0.385 \\ \hline
\multicolumn{4}{l}{\textbf{User Representation Learning}} \\ \hline
VGG16 & 0.163 & 0.558 & 0.163 \\
VGG16 + BERT & 0.186 & 0.749 & 0.188 \\ 
ResNet-50 & 0.194 & 0.725 & 0.195 \\ 
ResNet-50 + BERT  & \textbf{0.197} & \textbf{0.733} & \textbf{0.197} \\ 
\hline
\end{tabular}
%}
\end{center}
\end{table}

%% file: main_result.tex
\begin{table}[t]
\caption{Experimental results for user-challenge participation prediction task}
\label{tab:main}
\begin{center}
% \small
%  \resizebox{0.5\textwidth}{!}{
\begin{tabular}{c|ccc}
\hline
\textbf{Model} & \textbf{Macro-Prec} & \textbf{Macro-Rec} & \textbf{Macro-F1} \\ \hline \hline
VGG16 & 0.017 & 0.200 & 0.083 \\ 
VGG16 + BERT & 0.188 & 0.750 & 0.188 \\ 
ResNet-50 & 0.050 & 0.059 & 0.050 \\
ResNet-50 + BERT & 0.187 & 0.750 & 0.188 \\ 
\textsf{deepChallenger} & \textbf{0.494} & \textbf{0.933} & \textbf{0.494} \\
\hline
\end{tabular}
%}
\end{center}
\end{table}